\documentclass{article}

\usepackage{arxiv}

\usepackage[utf8]{inputenc} 
\usepackage[T1]{fontenc}    
\usepackage{hyperref}       
\usepackage{url}            
\usepackage{booktabs}       
\usepackage{amsfonts}       
\usepackage{nicefrac}       
\usepackage{microtype}      
\usepackage{lipsum}
\usepackage{graphicx}
\usepackage{gensymb}
\usepackage{adjustbox}
\usepackage{multirow}
\graphicspath{ {./images/} }

\title{Feasibility of assessing cognitive impairment via distributed camera network and privacy-preserving edge computing}

\author{
 Chaitra Hegde \\
  Department of Biomedical Informatics\\
  Emory University\\
  School of Electrical and Computer Engineering\\
  Georgia Institute of Technology \\
  Atlanta, GA\\
  \texttt{chegde@gatech.edu} \\
   \And
 Yashar Kiarashi \\
  Department of Biomedical Informatics\\
  Emory University\\
  Atlanta, GA \\
  \texttt{yash@dbmi.emory.edu} \\
  \And
 Allan I Levey \\
  Department of Neurology\\
  Emory University\\
  Atlanta, GA \\
  \And
  Amy D Rodriguez \\
  Department of Neurology\\
  Emory University\\
  Atlanta, GA \\
  \And
  Hyeokhyen Kwon$^{*}$ \\
  Department of Biomedical Informatics \\
  Emory University \\
  Department of Biomedical Engineering \\
  Georgia Institute of Technology \\
  Atlanta, GA \\
  \texttt{hyeokhyen.kwon@dbmi.emory.edu} \\
  \And
  Gari D Clifford\thanks{Joint senior authors.} \\
  Department of Biomedical Informatics \\
  Emory University \\
  Department of Biomedical Engineering \\
  Georgia Institute of Technology \\
  Atlanta, GA \\
  \texttt{gari@gatech.edu} \\
}

\begin{document}
\maketitle

\begin{abstract}

\noindent
{\normalsize \textbf{INTRODUCTION:} Mild cognitive impairment (MCI) is characterized by a decline in cognitive functions beyond typical age and education-related expectations. Since, MCI has been linked to reduced social interactions and increased aimless movements, we aimed to automate the capture of these behaviors to enhance longitudinal monitoring.

\noindent
\textbf{METHODS:} 
Using a privacy-preserving distributed camera network, we collected movement and social interaction data from groups of individuals with MCI undergoing therapy within a 1700$m^2$ space. 
We developed movement and social interaction features, which were then used to train a series of machine learning algorithms to distinguish between higher and lower cognitive functioning MCI groups.

\noindent
\textbf{RESULTS:} 
A Wilcoxon rank-sum test revealed statistically significant differences between high and low-functioning cohorts in features such as linear path length, walking speed, change in direction while walking, entropy of velocity and direction change, and number of group formations in the indoor space.
Despite lacking individual identifiers to associate with specific levels of MCI, a machine learning approach using the most significant features provided a 71\% accuracy.  

\noindent
\textbf{DISCUSSION:} We provide evidence to show that a privacy-preserving low-cost camera network using edge computing framework has the potential to distinguish between different levels of cognitive impairment from the movements and social interactions captured during group activities.}


\end{abstract} 
\section{Background}

Mild cognitive impairment (MCI) is a syndrome where cognitive decline exceeds age and education norms without significantly disrupting daily activities. Typically observed in individuals over the age of 65~\cite{gauthier2006mild}, MCI is characterized by impairment in memory, executive functions, attention, speed of processing, perceptual-motor abilities, and language~\cite{gauthier2006mild}, and often precedes dementia, with over half of those diagnosed with MCI progressing  within five years~\cite{gauthier2006mild}.  
U.S. projections estimate 15 million people with Alzheimer's dementia or MCI by 2060, up from 6.08 million in 2017.~\cite{brookmeyer2018forecasting}.
Timely MCI diagnosis allows patients and caregivers to gain insights and develop coping strategies while individuals still possess the capacity to do so independently. 
However, diagnosis and treatment are often delayed due to limited access to expert assessment or the dismissal of early symptoms as part of normal aging~\cite{Beason-Held18008}.

MCI and its potential progression to dementia are linked to lack of physical exercise and infrequent mental or social stimulation~\cite{world2019risk, ISAACSON20181663}. 
Considering that social relationships play a significant role in maintaining cognitive function during aging~\cite{soc_disengagement, 10.1093/geronb/60.6.P320}, the study of social interactions within the MCI population becomes paramount. Notable alterations in motor and gait patterns, including aspects like balance and coordination~\cite{mci_balance}, gait velocity reduction, increased stride variability, and changes in stride time and length~\cite{10.1159/000445831,10.1371/journal.pone.0099318,WAITE200589}, are also evident. These behavioral markers hold promise in identifying signs of cognitive decline in home or daily settings, potentially prompting individuals to seek professional diagnosis and guidance. 

Numerous studies have explored the use of machine learning to analyze gait and spatial navigation for prescreening MCI. Typically, these studies involved dual-task assessments, where individuals performed a mental load task, such as simple mathematics, while walking.  Gait features such as stride time, step time, single support time, swing time, double support time, stance time, stride length, and step length have been utilized in conjunction with classification models like support vector machines to distinguish between MCI, dementia, and healthy populations~\cite{9175955,ghoraani2021detection,mci_movement_complex}. 
However, the data for these studies were collected in controlled environment, which may not generalize well to real-world scenarios. Several studies have deployed motion sensors in the homes of elderly individuals living alone, equipping these residences with a range of unobtrusive sensors, including motion detectors and environmental sensors~\cite{10.3233/AIS-160385,dawadi2013automated,JAVED2021102572,HAYES2008395,RIBONI201657,7146521,mci_movement_complex}. These studies utilized various features such as room activity distribution~\cite{10.3233/AIS-160385}, an individual's ability to perform daily living activities~\cite{dawadi2013automated,JAVED2021102572}, walking speed, overall activity levels within the home~\cite{HAYES2008395}, and the time spent on different activities of daily living~\cite{RIBONI201657}. Due to the limitations of motion sensors, which can only detect the presence of individuals in a room, these studies were confined to single-resident households. The sensors were limited in their ability to understand spatial usage and navigation behaviors, only capturing data at the room level without granular details on the motion or activity within the room, or interactions with other individuals. 
Social interactions in the MCI population remain relatively understudied. Previous works have often relied on self-reports or caregiver accounts, which may introduce bias or overlook critical details due to the inherent limitations of recalling specific events. 

In this feasibility study, we introduce a passive sensing pipeline designed to capture detailed movement (at a resolution of 1 to 2 meters) and social interaction cues during group activities in an MCI population. Comprising a distributed camera network covering 1700$m^2$, data were captured  over fourteen months in an attempt to  distinguish between groups of individuals with varying cognitive function levels. In particular, the individuals with MCI were active in a space in which there were also healthy individuals, mimicking a real-world situation involving caregivers and family.
\section{Methods}

\begin{figure*}[t]
    \centering
    \includegraphics[width=\linewidth]{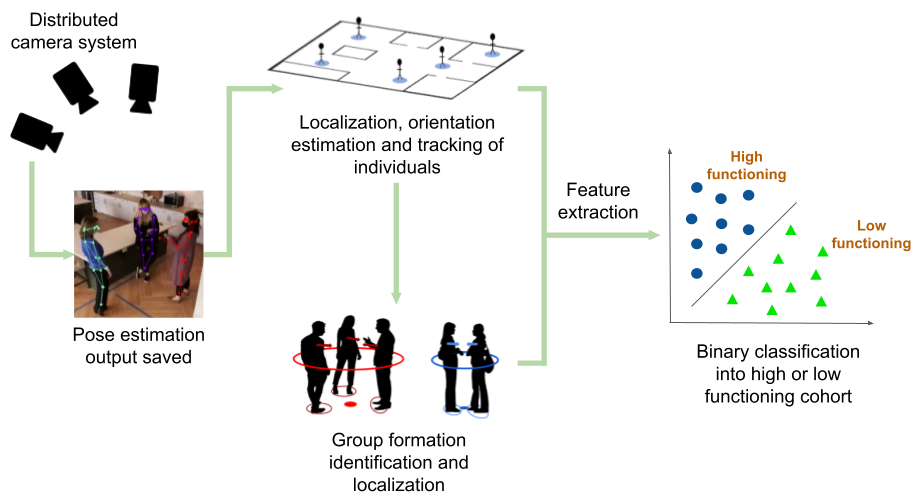}
    \caption{Proposed pipeline. A distributed camera network uses real-time pose estimation to collect privacy-preserving data from individuals in an indoor space. These keypoints are used to find the locations, orientations, and tracks of individuals in the indoor space~\cite{kwon2023indoor}. This is further used to identify and locate group formations~\cite{hegde2024_group}. Hand crafted features are extracted from the positions, orientations, tracks and group formation estimations. These features are used to classify a cohort as either a high- or low-functioning MCI cohort.
    }
    \label{fig:flow}
\end{figure*}

We employed a distributed camera system to gather behavioral data from MCI cohorts. Movement and social interaction features were extracted from this data and used with classifiers to categorize cohorts as low or high-functioning. The overall system is depicted in \autoref{fig:flow}.

\subsection{Data Collection}

\subsubsection{Study Site}
\begin{figure*}[t]
    \centering
    \includegraphics[width=\linewidth]{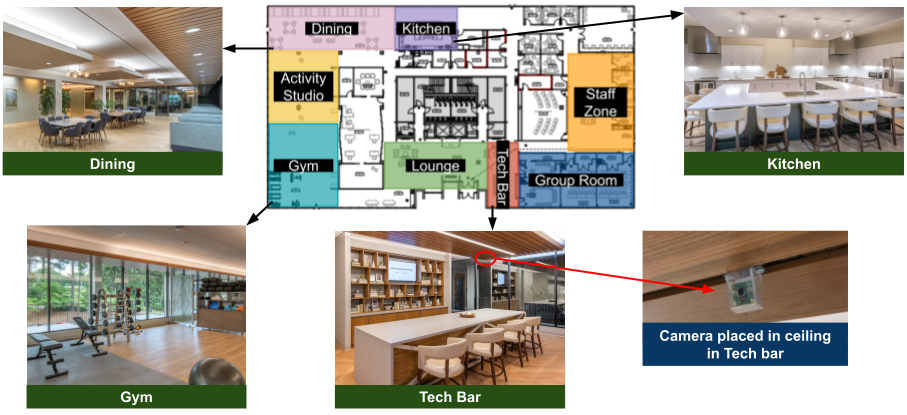}
    \caption{The layout of the indoor space used in this study, spanning 1,700$m^2$,  along with pictures from various functional areas within this space. Our study site has various regions to provide physical and cognitive training relating to activities in daily living for individuals with MCI. These areas include a gym, dining area, kitchen, lounge, activity area, tech bar, and staff zone. A picture of one of the cameras installed in the ceiling of the tech bar is also shown.
    }
    \label{fig:ep6_areas}
\end{figure*}

The data in this study was collected at the Charlie and Harriet Shaffer Cognitive Empowerment Program (CEP)\footnote{https://empowerment.emory.edu/}, a group therapeutic program and space designed for individuals with MCI.
In this program, participants engage in weekly activities such as cooking, exercise, and cognitive training from 9 am to 3 pm within a designated indoor space designed for therapeutic purposes as shown in \autoref{fig:ep6_areas}, spanning approximately 1,700$m^2$. Therapeutic staff and caregivers of the participants with MCI are also present to ensure their safety. 
Our analysis focuses solely on data from three daily breaks (two 15-minute, one 30-minute), when participants move and socially interact freely.

\subsubsection{Cognitive Assessment}
\begin{figure}[t]
    \centering
    \includegraphics[height=.7\linewidth,width=\linewidth]{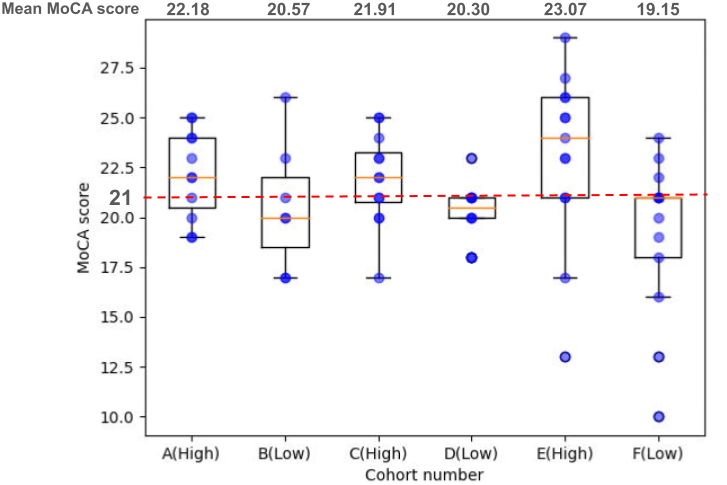}
    \vspace{-0.1in}
    \caption{Box plot illustrating the median, interquartile range, maximum and minimum of MoCA scores for each cohort. The MoCA scores for 66 subjects belonging to one of six cohorts are represented by the blue circles. Note that multiple individuals can have the same MoCA scores, resulting in overlapping blue circles, which appear as darker blue circles. The mean MoCA score for each cohort is displayed at the top of the plot. The threshold of 21 that is used to assign cohorts as either high functioning or low functioning is depicted by the dotted red line. The class that each cohort belongs to, i.e. high or low functioning, is denoted next to the cohort names.
    }
    \label{fig:moca_box_plot}
\end{figure}

Participants undergo the Montreal Cognitive Assessment (MoCA), a 30-point cognitive screening test evaluating various cognitive functions like short-term memory, visuospatial abilities, executive functioning, language, attention, and orientation~\cite{nasreddine2005montreal}. Scores greater than or equal to 26 indicate normal cognition, 18 to 25 suggest mild cognitive impairment, 10 to 17 imply moderate impairment, and less than 10 indicate severe impairment. Participants are allocated to a cohort with individuals of similar MoCA scores, thereby forming multiple cohorts representing high or low levels of cognitive functioning. 
\noindent We examined a total of 66 subjects with MoCA scores ranging from 10 to 29,  divided into six cohorts (A-F) containing 11, 7, 12, 10, 13, and 13 subjects respectively.
\autoref{fig:moca_box_plot} illustrates the distribution of MoCA scores within each cohort. Cohorts are categorized as either high functioning if the mean MoCA score exceeds 21 or low functioning if the mean MoCA score is 21 or below. This threshold selection of 21 is based on its positioning midway between the upper and lower limits of MoCA scores for MCI, namely 18 and 25. Our interpretation of high and low functioning cohorts identifies cohorts B, D and F as low functioning and cohorts A, C, and E as high functioning.

\subsection{Behavior Sensing Framework}

\subsubsection{Distributed Camera Network with On-device Pose Estimation}
\begin{figure}[t]
    \centering
    \includegraphics[width=\linewidth]{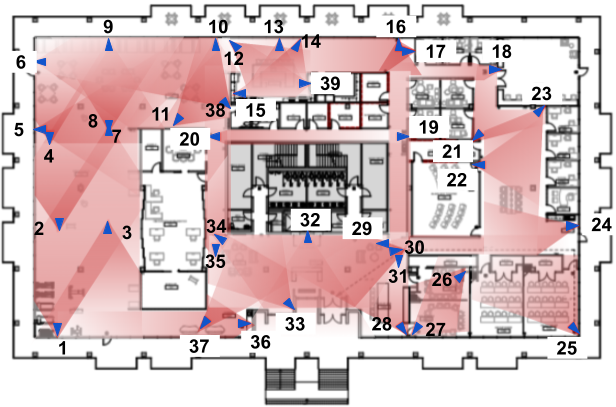}
    \vspace{-0.1in}
    \caption{Locations and coverage of the 39 edge computing camera systems deployed in the ceiling of the indoor space. The blue triangles depict the camera placement position and orientation. The cameras point in a direction away from the vertex, perpendicular to the short base, with a solid angle viewing region denoted by the shaded red regions. White areas are not covered by cameras, either by choice or infrastructural limitations.}
    \label{fig:camera_coverage_numbered}
\end{figure}

We installed 39 edge computing and camera devices, costing less than \$150 each, which were strategically deployed throughout the therapeutic space as shown in \autoref{fig:camera_coverage_numbered}. These devices consisted of a Raspberry Pi camera module V2 (Sony IMX219 8-megapixel sensor) and a Google Coral TPU USB Accelerator, integrated into a Raspberry Pi 4 model B to create an edge computing unit, running a multiperson 2-D pose estimation model~\cite{papandreou2018personlab} in real-time at 1 Hz. The pose estimation model detects the x and y coordinates of 17 keypoints across the body, such as knees, eyes, and nose,
for multiple people in the frame. Bounding boxes around people were also estimated from keypoint locations. 
The derived poses, along with human bounding box images, were stored for subsequent downstream analysis in an on-premise fog server. Raw video frames were discarded, ensuring that personally identifiable data was not stored and transmitted over the network, thereby upholding individual privacy. 
See Kwon \textit{et al.}~\cite{kwon2023indoor} for more details.

\subsubsection{Multi-view Multi-person Localization, Body Orientation Estimation, Tracking, and Group Localization}
\label{subsubsec:tracking_group_detection}
Following our previous work, the keypoints and bounding boxes were utilized to perform multi-view multi-person localization, body orientation estimation, and tracking analysis to capture the movements of individuals in the therapeutic space. We achieved an average localization error of 1.41 m, a multiple-object tracking accuracy score of 88.6\%, and a mean absolute body orientation error of 29$\degree$. More details about the procedure can be found in Kwon \textit{et. al.}~\cite{kwon2023indoor}. 
The estimated location and body orientation of individuals were further processed to detect group formations within the indoor environment using a clustering method. Groups of two or more individuals could be detected with an F1 score of up to 90\%. More details can be found in Hegde \textit{et al.}~\cite{hegde2024_group}. 

\subsection{Behavior Features for Group Activities in MCI }
\label{subsec:feature_analysis}

We extracted movement and social interaction features to highlight behavioral differences between cohorts with high and low cognitive functions, motivated by previous works~\cite{9175955,ghoraani2021detection,mci_social}. The features are explained below.

\subsubsection{Linear path lengths}
These values represent the lengths of uninterrupted walking paths without alterations in direction, determined by the individuals' positions . Previous research has suggested that linear path lengths (LPL) follow a Levy distribution when people walk in a pedestrian setting~\cite{murakami2019levy}. We hypothesized that the distribution of linear path lengths would differ between groups of varying cognitive functioning levels.
The feature calculations were based on the tracked positions of individuals.  
From the individual trajectories from our multi-person tracking, the detection of a change in walking direction was calculated by assessing the angle alteration between consecutive positions within the track. For instance, if we consider the consecutive three positions in a track denoted as $P_1=(x_1,y_1)$, $P_2=(x_2,y_2)$, and $P_3=(x_3,y_3)$, the angle change, $\theta$, was computed by analyzing the line segments formed by $\vec{P}_{1,2}= P_2-P_1$ and $\vec{P}_{3,2}=  P_2-P_3$, where 
$\theta = arccos(\frac{\vec{P}_{1,2}\cdot \vec{P}_{3,2}}{|\vec{P}_{1,2}|\cdot|\vec{P}_{3,2}|})$.
We considered a 20$\degree$ change in walking direction to constitute a new linear path, accounting for the noise in position and body orientation estimation. Consequently, consecutive lines with $\theta \leq 20\degree$ were considered part of a longer, uninterrupted linear path. 
From the collection of linear path lengths for all individuals in a cohort, we computed the mean and standard deviation of linear path lengths for each break duration. 

\subsubsection{Speed of walking}
Multiple studies have shown that walking speed decreases for people with cognitive impairment~\cite{9175955,ghoraani2021detection}.  
To determine the speed of walking for each detected individual, the linear path lengths described earlier were divided by the number of positions constituting each linear path, $S_i = LPL_i/n_i$, for $i$th LPL. The number of positions represented the time duration taken to traverse the linear path, given the 1Hz sampling rate in our passive sensing framework.  
The mean and standard deviation for the speeds of walking for all individuals in a cohort during a break period were used for classification. 

\subsubsection{Direction change}
We hypothesized that frequent changes in direction could be a sign of confusion and, hence, correlate to cognitive decline~\cite{kamil2021detection}. Specifically, we computed the mean and standard deviation of $\theta$ computed from LPL for all individuals' movement from a cohort observed during the break times. 

\subsubsection{Entropy of walking velocity}
Velocity entropy, a measure of complexity, was determined as the sample entropy of the speed computed between consecutive estimated locations, $V_t=|\vec{P}_{t,t+1}|$, where $\vec{P}_{t,t+1}$ represents the distance moved between two consecutive position samples. We hypothesize that walking velocity will change more dynamically in individuals with severe cognitive impairment due to the aimless movements associated with MCI~\cite{mci_movement_complex}.
During the break periods, we computed sample entropy from speed changes, $E_v = SampleEngropy(V_{1:T})$, where $T$ is the number of locations estimated for $v$th trajectory during the break session.
Finally, we extracted the mean and standard deviation of $E_{v=1,\cdots,K}$ from all individuals in a cohort for each break period.

\subsubsection{Entropy of orientation change}
We calculated the entropy from orientation changes $\Delta\theta_{t-1,t} = \theta_{t} - \theta_{t-1}$, between each timestep detected from the trajectory. This entropy of orientation change is denoted as $E_o=SampleEntropy(\Delta\theta_{1:T})$. 
Similar to velocity entropy, we derived the mean and standard deviation of $E_o$ from multiple trajectories available for each cohort within a break period.

\subsubsection{Levy distribution parameters for linear path lengths}
We hypothesized that the distribution of linear path lengths of healthy individuals follows a Levy distribution, while those of individuals with MCI exhibit more Brownian motions~\cite{hegde2022_socialdist, mci_movement_complex}. 
We fit a Levy distribution to the linear path lengths of each individual, extracting the location ($\mu$) and scale ($c$) parameters. The mean and standard deviation of these parameters for all individuals of a cohort, observed during the break sessions, were used for classification.

\subsubsection{Overall group formations}
Studies have shown that the total number of social interactions is lower in individuals with MCI from self-reports and surveys~\cite{mci_social}. 
We used the detected group formations as a proxy for social interactions. 
First, we counted the number of detected groups, $g_t$, and the number of individuals participating in group activities, $n_t$, at time $t$.
Then, for a cohort, we normalize the number of detected groups with the number of individuals participating in group activities, $G_t = g_t/n_t$.
This is to take account of the varying size and dynamics of group activities occurring over time.
Smaller $G_t$ can potentially mean that a large-sized group was detected ($g_t \downarrow$ \& $n_t \uparrow$), and larger $G_t$ can mean that multiple small-sized groups were detected ($g_t \uparrow$ \& $n_t \downarrow$).
Lastly, we derived mean and standard deviation of $G_{1:T}$ over each break duration for a cohort.

\subsubsection{Region-specific group formation}
As shown in \autoref{fig:ep6_areas}, the therapeutic space used in this study includes several regions dedicated to different activities such as cooking, dining, exercise, and classroom sessions. 
We hypothesized that during break times, groups with different levels of cognitive functioning would use these spaces differently.
At a specific region of the therapeutic space, $r$, at time $t$, we computed the number of groups formed in distinct regions of the facility, $g^r_t$, which was normalized by the number of individuals participating in group activities in that region, $n^r_t$.
The normalized region-specific group formation, $G^r_t = g^r_t/n^r_t$, was collected over each break period for each cohort, and the mean and standard deviation for each region was derived. \\

Overall, the movement features included linear path lengths, walking speed, direction change, walking velocity entropy, orientation change entropy, and Levy distribution parameters. The social interaction features encompassed overall group formations and region-specific group formations.

\subsection{Cohort-level Classification of Cognitive Impairment}
The derived features were used for binary classification of high and low levels of cognitive function of each cohort according to \autoref{fig:moca_box_plot}.
Before model training and testing, the behavior features were rescaled to lie between 0 and 1 to account for the different ranges of values in each feature. 
The rescaled features were then fed to classification models, such as Support Vector Machines (SVM).
\section{Experiments}

315 break sessions (15-30 minutes each) were recorded over fourteen months across six cohorts. Each break session served as a single sample for deriving behavior features. Based on the mean MoCA score of each cohort, they were labeled as either high (168 instances) or low (147 instances) cognitive functioning, serving as the gold-standard measures.
For classification, we employed radial basis function-based non-linear support vector machine (SVM), XGBoost (XGB), logistic regression (LR), and Lasso binary classification algorithms in a leave-one-sample-out cross-validation manner.
Model performance was compared using all features, only movement features, and only social interaction features, evaluated by precision, recall, F1 score, and accuracy.
The Wilcoxon Rank Sum Test assessed statistical differences in raw feature distributions between high and low-functioning cohorts. The null hypothesis stated that the feature distributions for both groups were similar, with a significance threshold set at $p < 0.05$. It's important to note that the features examined in this test were raw features' distributions, not their means and standard deviations.
\section{Results}

\begin{table*}[t]
    \centering
    \caption{Precision, recall, F1 score and accuracy along with 95\% confidence intervals when SVM, XGBoost, logistic regression and lasso binary classifications were used with all the features, only social interaction features and only movement features.
    (\textbf{Bold} shows the best performance.)}
    \begin{adjustbox}{width=\textwidth}
    \begin{tabular}{|c|c||c|c|c|c|}
    \hline
        \textbf{Metrics} & & \textbf{Precision} & \textbf{Recall} & \textbf{F1 Score} & \textbf{Accuracy}\\
        \hline
        \multirow {4}{*}{All features} & SVM & $0.62\pm0.054$ & \textbf{0.72$\pm$0.050} & $0.67\pm0.052$ & $0.66\pm0.052$ \\
          & XGB & \textbf{0.69$\pm$0.051} & $0.66\pm0.052$ & \textbf{0.68$\pm$0.052} & \textbf{0.71$\pm$0.050} \\
          & LR & $0.65\pm0.053$ & $0.65\pm0.053$ & $0.65\pm0.053$ & $0.68\pm0.052$ \\
          & Lasso & $0.67\pm0.052$ & $0.64\pm0.053$ & $0.65\pm0.053$ & $0.68\pm0.051$ \\
        \hline 
          
        \multirow {4}{*}{Social interaction features} & SVM & $0.63\pm0.053$ & $0.71\pm0.050$ & $0.67\pm0.052$ & $0.67\pm0.052$ \\
         & XGB & $0.64\pm0.053$ & $0.64\pm0.053$ & $0.64\pm0.053$ & $0.67\pm0.052$ \\
         & LR & $0.62\pm0.054$ & $0.62\pm0.054$ & $0.62\pm0.054$ & $0.64\pm0.053$ \\
         & Lasso & $0.63\pm0.053$ & $0.63\pm0.053$ & $0.63\pm0.053$ & $0.66\pm0.052$ \\
         \hline

        \multirow {4}{*}{Movement features} & SVM & $0.64\pm0.053$ & $0.70\pm0.051$ & $0.67\pm0.052$ & $0.68\pm0.052$ \\
         & XGB & $0.64\pm0.053$ & $0.59\pm0.054$ & $0.61\pm0.054$ & $0.65\pm0.053$ \\
         & LR & $0.65\pm0.053$ & $0.57\pm0.055$ & $0.60\pm0.054$ & $0.65\pm0.053$ \\
         & Lasso & $0.66\pm0.052$ & $0.59\pm0.054$ & $0.62\pm0.054$ & $0.66\pm0.052$ \\
        
        \hline
    \end{tabular}
    \end{adjustbox}
    \label{tab:moca_classification}
\end{table*}
\autoref{tab:moca_classification} shows classification results for SVM, XGBoost, logistic regression, and lasso binary classifiers using different feature sets. XGBoost achieved the highest F1 score (0.68) using all features, slightly outperforming SVM (0.67). The SVM demonstrated the highest performance and stability across all feature combinations. Logistic regression and lasso binary classifier both reached 0.65 F1 score with all features, outperforming their scores with social interaction features by 3\% and 2\%, and movement features by 5\% and 3\%, respectively. XGBoost's performance decreased by 4\% with only social interaction features and 7\% with only movement features.

\begin{table}
    \centering
    \caption{P-values of the Wilcoxon rank sum test performed on the features for distinguishing high and low-functioning cohorts. 
    The p-values with a star (*) indicate features with statistically significant differences ($p<0.05$) in distributions between low and high-functioning groups.}
    \begin{tabular}{|c|c|}
        \hline
        \textbf{Feature name} & \textbf{P-value}\\
        \hline
        Linear path length & < 0.0001*\\ 
        \hline
        Walking speed & < 0.0001*\\ 
        \hline
        Direction change & < 0.0001*\\ 
        \hline
        Velocity entropy & < 0.0001*\\
        \hline
        Orientation change entropy & < 0.0001*\\
        \hline
        Levy $\mu$ parameter & 0.0676\\
        \hline
        Levy $c$ parameter & 0.0364*\\%
        \hline
        Number of groups in therapeutic space & < 0.0001* \\
        \hline
    \end{tabular}
    \label{tab:ranksum}
\end{table}
In \autoref{tab:ranksum}, the p-values for each tested feature are presented to assess their statistical significance in differentiating cohorts based on cognitive function levels.
All features except ``Levy $\mu$ parameter'' demonstrated statistical significance ($p < 0.05$). 
\section{Discussion}
\label{sec:discussion}

\subsection{Classifying High and Low functioning Cohorts}
\begin{figure*}[hp]
    \centering
    \begin{tabular}{c}
    \includegraphics[width=0.9\textwidth]{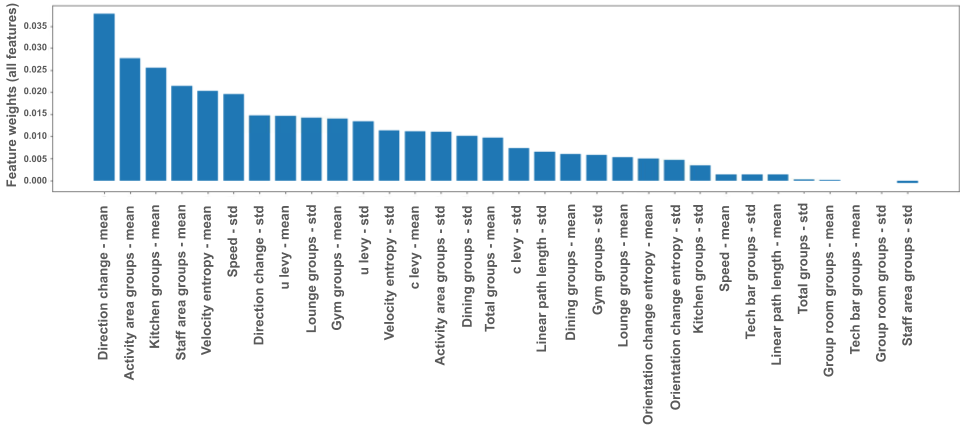} \\
    \includegraphics[width=0.9\textwidth]{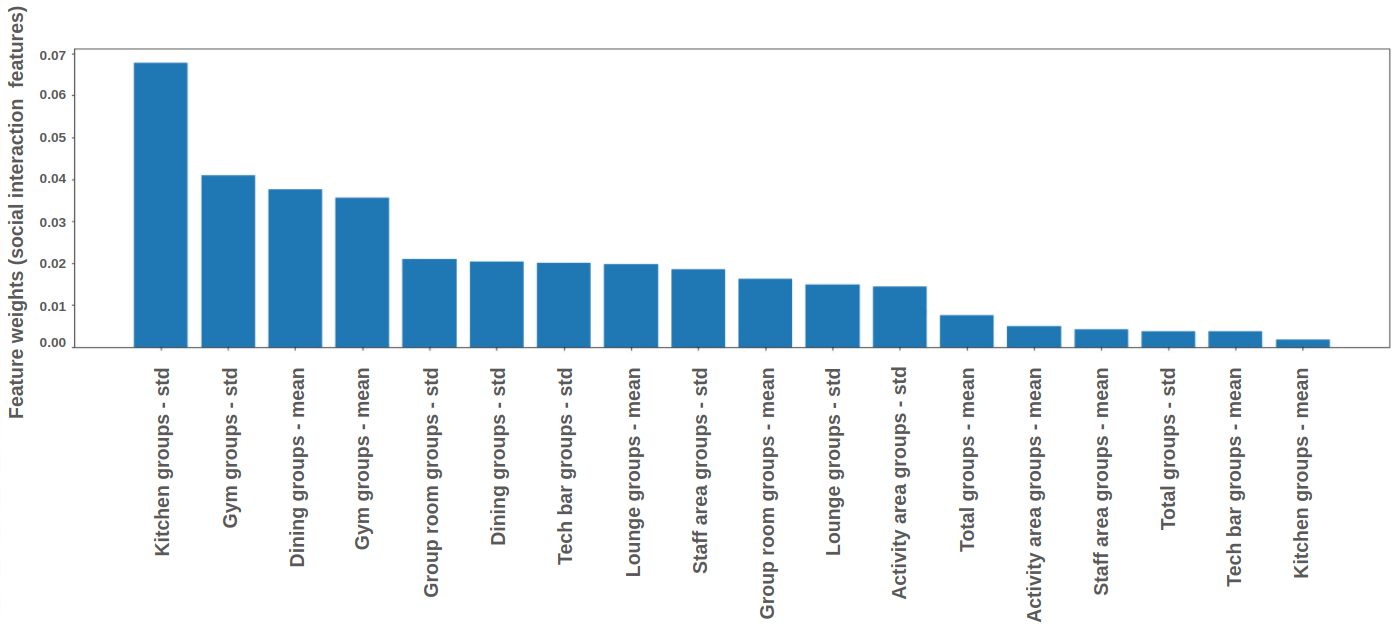} \\
    \includegraphics[width=0.9\textwidth]{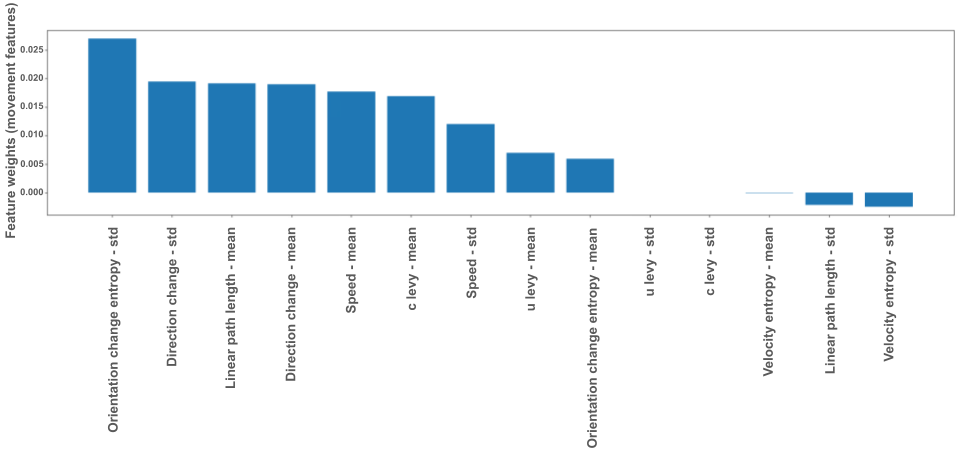} 
    \end{tabular}
    \caption{Feature importance analysis for the SVM models used for classification. The top figure shows the feature weights when using all features, the middle figure shows the feature weights when only using social interaction features, and the bottom figure shows the feature weights when only using movement features.}
    \label{fig:feature_importance}
\end{figure*}

\autoref{tab:moca_classification} shows that the SVM slightly outperforms the logistic regression and lasso binary classifiers in terms of F1 score across all feature combinations, with the difference being most pronounced with movement features, and surpasses XGBoost when only social interaction or movement features are used. Notably, the highest F1 scores for XGBoost, logistic regression and lasso binary classifier occur when all features are used, followed by social interaction features, and then movement features. In contrast, the SVM maintains consistent performance across all feature combinations.
These results suggest that the most informative results are obtained when all features are used. Precision remains relatively stable across all feature combinations for SVM, logistic regression and lasso binary classifier. XGBoost achieves the highest precision with all features, but it decreases by 5\% when only social interaction and movement features are used. Recall drops significantly for XGBoost, logistic regression and lasso binary classifier when only movement features are used, reducing their F1 scores. This suggests a higher rate of missed "low functioning" samples (false negatives) with this feature combination. The dataset includes 14 movement features and 18 social interaction features (see \autoref{fig:feature_importance} for features) which likely explains the performance degradation when only movement features are used due to the reduced information from fewer features. The SVM, being a more complex and non-linear model, may effectively differentiate between classes using only movement or social interaction features, thus showing little improvement when all features are combined.

\subsection{Feature Analysis}
The outcomes of the Wilcoxon rank sum test, as presented in \autoref{tab:ranksum} indicate statistically significant differences in the distributions of most features between high and low-functioning MCI cohorts, suggesting the potential predictive ability of these features. Specifically, the walking speed of lower-functioning groups was lower compared to higher-functioning groups, especially at higher speeds. This aligns with previous research that has found a decrease in walking speed, particularly at higher speeds, to be associated with cognitive impairment~\cite{knapstad_walkingSpeed, hackett_walkingSpeed}. Furthermore, the mean linear path lengths were slightly longer in lower-functioning cohorts  ($\approx4.9m$) compared to higher-functioning cohorts ($\approx4.5m$), suggesting that higher-functioning cohorts engaged in more directed and intentional movement. Additionally, the velocity and orientation change entropies were greater in the lower-functioning cohorts, suggesting more confusion during movement in these groups. Higher numbers of group formations were more commonly found in higher-functioning cohorts than in lower-functioning cohorts. Notably, the feature ``Levy $\mu$ parameter'' had $p>0.05$, indicating comparable distributions across low and high-functioning cohorts. This observation may be attributed to the inclusion of individuals with normal cognition (such as care partners and therapeutic staff) alongside those with MCI in our cohort data. This mixed composition could have impacted the distributions of the Levy features, potentially diluting their discriminatory power. Most of these features, other than walking speed, have not been previously studied in detail and could potentially serve as new parameters to assess MCI.

\autoref{fig:feature_importance} presents the feature importance obtained using the permutation feature importance method for the SVM for three feature combinations: using all features (top), only social interaction features (middle), and only movement features (bottom). These plots highlight the features that contributed most to the classification of high- and low-cognitive functioning cohorts by the SVM models, emphasizing the relative importance of each feature within the model. 
When using all features (i.e., social interaction and movement features, \autoref{fig:feature_importance}, top), the SVM model identified the most important features as mean and standard deviation of direction change, mean and standard deviation of velocity entropy, standard deviation of walking speed, mean and standard deviation of location parameter of the Levy distribution, mean number of group formations in the activity area, kitchen, staff area and gym and the standard deviation of number of group formations in the lounge. The lounge, gym and activity areas are specifically designed to encourage social interactions, providing open spaces and seating. It is interesting to observe that the Levy location parameters was among the most important features when all features were used as well as when only movement features were used, contrary to Wilcoxon rank sum analysis. We suspect that the Levy parameters provide useful signals when combined with other movement features, such as walking speed and orientation changes, helping to find better classification margins for the SVM. We also hypothesize that Levy parameters will provide useful information when the analysis is done on an individual level, which is part of our future work.
When only considering social interaction features (\autoref{fig:feature_importance}, middle), the most important features were the mean number of groups in the dining area and gym, and the standard deviation of the number of groups in the kitchen and gym.  
When only considering movement features (\autoref{fig:feature_importance}, bottom), the most useful features were mean linear path length, direction change, walking speed and scale parameter of the Levy distribution and standard deviation of orientation change entropy and direction change. Previous studies have linked aimless movements and frequent angle changes to MCI. Our findings, which show that the mean and standard deviation of direction change significantly contributed to classification when using only movement features and all features, are consistent with these studies~\cite{mci_movement_complex,wang2024dual,kamil2021detection}.

\subsection{Limitation \& Future Direction}
Our study uniquely focuses on group analysis of cohorts with varying cognitive impairment levels coexisting with healthy individuals in a real-world setting. Despite higher noise in group-based data, we achieved a 68\% F1 score in distinguishing low- and high-functioning cohorts. 
This lays the foundation for further exploration to validate the robustness of the proposed passive monitoring systems for MCI.

We envision that future iterations of this system could be implemented in nursing homes or hospitals to monitor behavioral pattern changes over time for older adults with cognitive impairments. Our future work will focus on distinguishing subjects with MCI from individuals with normal cognition by integrating Bluetooth and video-based tracking for detailed, individual-level analysis of cognitive impairment. In a previous study, we developed a Bluetooth-based indoor localization and tracking system that utilized Bluetooth sensors carried by subjects with MCI, achieving a localization error of $4.4m$~\cite{kiarashi2023graph}. We are now developing a sensor fusion framework that combines camera and Bluetooth sensors. Additionally, we will conduct an in-depth study of Levy parameters for individuals with MCI and normal cognition. These parameters have proven useful for distinguishing movement patterns in cohorts with high and low cognitive functions, despite overlapping feature distributions and the presence of healthy individuals. We also plan to apply temporal deep learning methods, such as recurrent neural networks~\cite{lalapura2021recurrent}, to model longitudinal changes in movement and group activity behavior related to cognitive impairment.
\section{Conclusion}

In an aging society, passively monitoring the movements and social interactions of older adults is crucial for objectively quantifying cognitive impairments. Previous studies have typically examined such behaviors in controlled lab environments or through surveys~\cite{mci_movement_complex,HAYES2008395,mci_social}.
Our work demonstrated that a distributed camera system, installed in a large therapeutic environment spanning 1700$m^2$, can effectively quantify longitudinal (fourteen-month) group activity and movements during break times for cohorts with MCI. We also showed that movement and social interaction based features, such as linear path length, walking speed, direction change, velocity and direction change entropy, Levy parameters and number of group formations have discriminative ability, consistent with previous studies~\cite{mci_movement_complex,HAYES2008395,mci_social}. This is even true in noisy settings where healthy individuals are present alongside those with MCI and where individual identifiers are missing, reflecting real-world scenarios with privacy constraints. This approach helps move beyond controlled tests and questionnaires to assess the cognitive abilities of older adults, potentially identifying early signs of dementia through longitudinal passive monitoring.
\section*{Acknowledgment}
HK, GC, and AL are partially funded by the National Institute on Deafness and Other Communication Disorders (grant \# 1R21DC021029-01A1). The Cognitive Empowerment Program is supported by a generous investment from the James M. Cox Foundation and Cox Enterprises, Inc., in support of Emory’s Brain Health Center and Georgia Institute of Technology. GC is partially supported by the National Center for Advancing Translational Sciences of the National Institutes of Health under Award Number UL1TR002378.

Declarations of interest: none.


\bibliographystyle{unsrt}  
\bibliography{references}  

\end{document}